\documentclass[letterpaper, 10 pt, conference]{ieeeconf}  

\usepackage{amsthm}
\usepackage{times}
\usepackage{hyperref}
\usepackage{textcomp}
\usepackage{graphicx}
\usepackage{comment}
\usepackage{amsmath,amsfonts,amssymb,color}

\newtheorem{proposition}{Proposition}
\newtheorem{lemma}{Lemma}
\newtheorem{theorem}{Theorem}
\newtheorem{remark}{Remark}
\newtheorem{example}{Example}

\newtheorem{problem}{Problem}
\renewcommand{\vspace}[1]{}

\newcommand{\bshe}[1]{\textcolor{black}{#1}}
\newcommand{\baike}[1]{\textcolor{black}{#1}}
\newcommand{\final}[1]{\textcolor{black}{#1}}
\usepackage{stackengine}
\def\cross{%
  \stackon[1ex]{\rule{0.4pt}{1.5ex}}{\rule{.75ex}{0.4pt}}}

\DeclareMathOperator*{\diag}{\textnormal{diag}}

\IEEEoverridecommandlockouts                              

\title{\LARGE \bf
Modeling Epidemic Spread: A Gaussian Process Regression Approach}
\author{Baike She$^{1},$ Lei Xin$^{2,\cross}, $ Philip E. Par\'e$^{3},$ Matthew Hale$^{1, *}$ 
\thanks{*This work is supported by the grant NSF-ECCS \#2238388 and by DARPA under grant HR00112220038.}
\thanks{$^{\cross}$ Corresponding Author.}
\thanks{$^{1}$Baike She and Matthew Hale are with the School of Electrical and Computer Engineering,
        Georgia Institute of Technology, Atlanta, GA, 30318, USA.
        {\tt\small bshe6@gatech.edu; matthale@gatech.edu}}%
\thanks{$^{2}$Lei Xin  is with the Department of Computer Science and Engineering, The Chinese University of Hong Kong, Hong Kong, 999077, China.
        {\tt\small leixin@cuhk.edu.hk}}%
\thanks{$^{3}$Philip E. Par\'e is with the Elmore Family School of Electrical and Computer Engineering, Purdue University, West Lafayette, IN 47907, USA.
        {\tt\small philpare@purdue.edu}}%
}
\begin{document}
\maketitle

\begin{abstract}
Modeling epidemic spread is critical for informing policy decisions aimed at mitigation. Accordingly, in this work we present a new data-driven method based on Gaussian process regression (GPR) to model epidemic spread \final{through the difference on the  logarithmic scale of the infected cases.} 
We bound the variance of the predictions made by GPR, 
which quantifies the impact of epidemic data on the proposed model. Next, we derive a high-probability error bound on the prediction error 
\baike{in terms of the distance between the training points and a testing point, the posterior variance, and the level of change in the spreading process,} and we assess how the characteristics of the epidemic spread and infection data influence this error bound. 
We present  examples that use GPR to model and predict epidemic spread by using real-world infection data gathered
in the UK during the COVID-19 epidemic. 
\baike{These examples illustrate that, under typical conditions, the prediction for the next twenty days has \baike{$94.29$\%} of the noisy data located within the $95$\% confidence interval, 
validating these predictions.} \final{We further compare the modeling and prediction results with other methods, such as polynomial regression, $k$-nearest neighbors (KNN) regression, and neural networks, to demonstrate the benefits of leveraging GPR in disease spread modeling.}
\end{abstract}
\section{INTRODUCTION}
Modeling and predicting the spread of diseases is critical for understanding spreading patterns and decision-making for epidemic mitigation~\cite{giordano2020modelling,woolhouse2011make}. Existing epidemic modeling and prediction techniques typically construct compartmental models by selecting model structures and parameters to fit spreading data~\cite{she2022mpcepi, she2022optimal}, e.g., in
the susceptible-infected-recovered (SIR) model. 
Distinct from existing works, we leverage Gaussian process regression to model  spreading trends by studying the number of infected cases directly, without using  any particular compartmental model. 


Gaussian process regression excels at capturing complex, nonlinear relationships without relying on predefined functional forms and can effectively handle small datasets, which are common in measuring 
disease spreading~\cite{senanayake2016predicting, velasquez2020forecast, zimmer2020influenza}.
\final{For instance,~\cite{senanayake2016predicting} employs a spatio-temporal covariance function and data from various states and all weeks of the year to model influenza-like illness forecasting. Meanwhile,~\cite{zimmer2020influenza} trains individual Gaussian process (GP) models for each forecast based on a relatively small set of features from previous weeks, resulting in small, but reliable prediction intervals.} \final{Moreover, although existing works leverage Gaussian processes to model daily infected cases, they often overlook the fact that both the daily infected cases and the associated noise cannot follow a normal or log-normal distribution~\cite{gonccalves2021covid}. Therefore, applying Gaussian process regression directly to daily infected cases may not produce accurate results. In addition, these studies typically focus on empirical results of applying GPR to model and predict disease
dynamics in time-series data, without delving into theoretical aspects.}


To address these challenges, we introduce an approach to modeling~\final{the change 
on the  logarithmic scale of the number of} infected cases using Gaussian process regression, while also providing insights into model uncertainty. We propose an upper bound on posterior variance to assess the impact of epidemic data and develop a high-probability error bound to examine how epidemic spread and infection data influence the accuracy of predictions and confidence in them. These results help bridge the gap between theoretical analyses and practical applications in epidemic modeling, paving the way for predictive control methods in future efforts. 

To illustrate this framework, we apply it to real COVID-19 infection data from the United Kingdom. \baike{These results show that by selecting appropriate parameters in the modeling process, 
%
predictions for twenty days into the future capture $94.29$\% of the actual data from those days within a $95$\% confidence interval,
which validates the prediction accuracy in practice. 
As we show, most prediction errors arise from either drastic changes in the spreading trend or a limited number of available data samples, both of which increase uncertainty.} \final{We further use the same dataset to compare the modeling and prediction results with other methods, such as polynomial regression, $k$-nearest neighbors (KNN) regression, and neural networks, to demonstrate the benefits of leveraging GPR for disease spread modeling.}

The rest of the paper is organized as follows: Section~\ref{sec_background} provides background and problem statements; Section~\ref{sec:gprspread} proposes the model, analyzes prediction uncertainty, and
illustrates the results with examples;  Section~\ref{Sec_Sen_analysis} presents sensitivity analyses on the impact of the data preprocessing process on the modeling and prediction results.
Section~\ref{sec_conclusion} concludes. 

\noindent\textbf{Notation: }
We use $\mathbb{R}$  and $\mathbb{N}$ to denote the sets of reals and naturals, respectively. 
We define $\underline{n} = \{1, 2,\dots, n\}$ for $ n\in \mathbb{N}_{>0}$. \bshe{
We use~$|S|$ to denote the cardinality
of a finite set~$S$.} A compact interval $\mathbb{T}$ is  of the form $[a,b]$, where $a, b\in\mathbb{R}_{\geq 0}$ and $a<b$.
Its length is given by $\hat{\mathbb{T}}=b-a$.
For $v\in\mathbb{R}^n$, we use $\diag\{v\}\in\mathbb{R}^{n\times n}$ for the diagonal matrix whose $i^{th}$ diagonal entry is $v_i$, $i\in\underline n$.
For a real symmetric  matrix $A\in\mathbb{R}^{n\times n}$, let $[A]_{ij}$ denote its $i^{th}j^{th}$ entry and 
$\lambda_{\max}(A)$ denote its largest eigenvalue. We write $I_n\in\mathbb R^{n\times n}$ for the identity matrix.
Let $\mathcal{N}(\mu,\sigma^2)$  denote the one-dimensional normal distribution with mean $\mu$ and variance $\sigma^2$. We use $\exp (\cdot)$ and $\log(\cdot)$ to denote the exponential function and \baike{the natural logarithmic function}, respectively. We use $p(\cdot)$ to represent the probability distribution of a random variable.
\section{Background and Problem Formulation}
\label{sec_background}
This section introduces Gaussian process regression and states the problems that we solve in the rest of the paper.
\subsection{Gaussian process regression}
\label{Sec: GPR}
We briefly introduce one-dimensional Gaussian process regression~\cite{williams2006gaussian}.
Consider an unknown function $f : \mathbb{R}\rightarrow \mathbb{R}$ and $n$ inputs captured by $X=[x_1,\dots, x_n]^{\top}\in\mathbb{R}^{n}$, where 
$x_i\in\mathbb{R}$, $i\in\underline n$, 
and $n\in\mathbb{N}_{>0}$. The 
corresponding outputs  are given by the vector
$F = [f(x_1),\dots, f(x_n)]^{\top} \in \mathbb{R}^{n}$, where the $n$ outputs in $F$
follow a joint Gaussian distribution. 
The mean of the distribution is given by ${m(X)=[m(x_1),\cdots,m(x_n)]^\top\in\mathbb{R}^n}$. 
Suppose that the observation of each output  $f(x_i)$ is corrupted with zero-mean independent Gaussian noise, i.e., $y(x_i) = f(x_i)+\varepsilon_i$,
\baike{where $\varepsilon_i \sim \mathcal{N}(0,\sigma_i^2)$, 
and $\sigma_i^2>0$ denotes the variance of $\varepsilon_i$.}
Then, the covariance matrix of the noise is 
\final{$\Sigma = \diag\{\sigma^2_1,\dots, \sigma^2_n\} \in\mathbb{R}^{n\times n}$}.
Using the noisy training dataset $\{(x_i,y(x_i))\}_{i=1}^{n}$, we can employ GPR to model the input-output relation $f:\mathbb{R}\rightarrow\mathbb{R}$ at a training location $x_i$, for $i\in\underline{n}$, as well as to 
predict the output $f(x^*)\in\mathbb{R}$ at some testing location $x^*\in\mathbb{R}$, where~$x^*$ is not necessarily
one of the~$x_i$'s.

Gaussian process regression is a kernel-based approach. Hence, we use $k(\cdot,\cdot):\mathbb{R}\times \mathbb{R}\rightarrow \mathbb{R}_{\geq0}$ to represent the potential kernel function~\cite{genton2001classes}. 
Let $K(X,X)\in\mathbb{R}^{n\times n}_{\geq0}$ denote the kernel matrix of the training points, where $[K(X,X)]_{ij}=k(x_i,x_j)$ 
denotes the covariance between two training points $x_i$ and $x_j$, for $i,j\in\underline{n}$. 
For a testing point~$x^*$, 
we define  $K(x^*,X)\in \mathbb{R}_{\geq 0}^{1\times n}$ as the kernel vector such that $[K(x^*,X)]_j=k(x^*,x_j)$, for $j\in\underline{n}$.
Therefore, $K(x^*,X)$ captures the covariances between the testing point  $x^*$ and all training points.
As Gaussian process regression operates as a Bayesian inference approach, 
we consider a zero-mean prior for generality~\cite{williams2006gaussian}, though
the results we develop can be generalized
for any other prior.

\bshe{Consider the posterior distribution for the predicted random variable $f(x^*)$ at the testing location $x^*$ conditioned on the noisy training data $\{(x_i,y(x_i))\}_{i=1}^{n}$.}
The posterior mean $m(x^*)$ and posterior variance $\sigma^2(x^*)$ at the testing location $x^*$ are given by the following result~\cite[Equation~(2.22)]{williams2006gaussian}.
\begin{proposition}
\label{Prop_GPR}
Let $Y=[y(x_1),\cdots, y(x_n)]^\top$. Then
we have $p(f(x^*) \mid \final{\{y(x_i)\}_{i=1}^n}) = \mathcal{N}(m_Y(x^*),\sigma_Y^2(x^*))$,
where
\begin{align*}
   m_Y(x^*) &= K(x^*,X)(K(X,X)+\Sigma)^{-1}Y,\\
   \sigma_Y^2(x^*) &= k(x^*,x^*) \\
   &\qquad- K(x^*,X)(K(X,X)+\Sigma)^{-1}K(x^*,X)^{\top}.
\end{align*}
\end{proposition}
\subsection{Problem Formulation} \label{Sec_PF}
\final{As discussed in the introduction, critical metrics such as the number of infected cases are essential for assessing epidemic spread. However, both these cases and their associated noise do not follow a normal or log-normal distribution~\cite{gonccalves2021covid}.
Furthermore, epidemic severity is typically monitored through population testing and data reporting, where observation noise is unavoidable. However, existing works often overlook these limitations and use Gaussian processes to model the spread by directly focusing on the trend of time-series infection cases. Additionally, missing cases frequently arise due to insufficient testing or underreporting.} Therefore, in this work, we solve the following problems:

\begin{problem} \label{prob:GPR}
\baike{Develop a new model that uses Gaussian process regression to model epidemic spread, and show the effectiveness of this approach analytically
and numerically.}
\end{problem}

\begin{problem} \label{prob:data}
\baike{Quantify how noise and data sample size affect the prediction results when using the model from Problem~\ref{prob:GPR}.} 
\end{problem}
\begin{problem} \label{prob:error}
\baike{Develop a high-probability error bound on the prediction error to analyze the impact of data, and validate this result on real-world data. 
}
\end{problem}
\begin{problem} \label{prob:comparions}
\final{Compare the GPR modeling and prediction results with other methods to demonstrate its benefits using real-world spread data.
}
\end{problem}
\section{Gaussian process regression for Modeling and Predicting Epidemic Spread} \label{sec:gprspread}
In Section~\ref{Sec_Model_Epidemic}, we model 
epidemic spread using Gaussian process regression, and then
quantify its accuracy in a lemma and an example. This solves
Problem~\ref{prob:GPR}. 
In Section~\ref{Sec_Epi_Prediction_Var}, we establish an upper bound on the prediction variance to assess the impact of infection data, 
which solves Problem~\ref{prob:data}.
Section~\ref{Sec_Epi_Prediction_Error} presents a high-probability error bound, explaining the relationship between the spreading dynamics, data, and prediction error, solving Problem~\ref{prob:error}. \final{We compare GPR with other methods, through Examples~\ref{Example_1} and~\ref{Example_3}, addressing Problem~\ref{prob:comparions}.}

\subsection{Modeling the Spread Using Gaussian Process Regression}
\label{Sec_Model_Epidemic}
We first solve Problem~\ref{prob:GPR} 
by introducing a model to capture disease spreading trends.
For an epidemic spreading process, 
we use $I(t)$ to denote a noisy observation of
the number of the infected cases  at time step $t\geq0$, i.e., $I(t)$ equals the number of infected cases at time~$t$ plus some noise. \bshe{We use $\bar{I}(t)$ to denote the true infected cases without observation noise.}
Consider the datasets $\{I(t_1), I(t_2), \dots, I(t_n)\}$ and $\{I(t_1 - \eta), I(t_2 - \eta), \dots, I(t_n - \eta)\}$ of an epidemic spreading process, measured at times $\{t_1, t_2, \dots, t_n\}$ and $\{t_1 - \eta, t_2 - \eta, \dots, t_n - \eta\}$, respectively, where $\eta > 0$, $t - \eta > 0$, and $n \in \mathbb{N}_{>0}$.
We assume that the change in the logarithmic scale of the number of infected cases between consecutive time steps approximates a Gaussian process. 
Thus, 
for $i\in\underline n$,
we define
\begin{align}
\label{eq:state}
\Delta(t_i) 
=\underbrace{\log (\bar{I} (t_i)) - \log (\bar{I}(t_{i}-\eta))}_{\bar{\Delta}(t_i)}+\varepsilon(t_i).
\end{align}
\begin{remark}
\final{
We mentioned in Section~\ref{sec_background} that infected cases typically do not follow a Gaussian or even a log-normal distribution. As discussed in~\cite{bhatt2023semi}, both the cases and the noise are often better modeled using Poisson or negative binomial distributions. The phenomenon where differences of log-transformed variables can approximate a Gaussian distribution is rooted in statistical principles such as variance stabilization~\cite{shao2008mathematical}, the Central Limit Theorem, and the Delta Method~\cite{oehlert1992note}. While the log-transformation of Poisson-distributed data may introduce skewness, differencing tends to normalize highly-skewed distributions. This behavior is often leveraged in fields like epidemiology to model relative changes in data empirically, including changes in disease counts~\cite{bland1996statistics, bosse2023scoring, Epinow2}. For example, studies like~\cite{bland1996statistics} and~\cite{bosse2023scoring} confirm that modeling the differences on the logarithmic scale of case counts improves predictability and aligns with Gaussian-based inference methods. Widely-used tools (e.g.,~\cite{Epinow2} for COVID-19 modeling) rely on relative changes in case counts, often log-differenced, to ensure statistical validity under Gaussian assumptions.
Therefore, we provide an analysis based on the first-order differences on the logarithmic scale of infected cases in~\eqref{eq:state}.}
\end{remark}
\begin{proposition}
\label{Prop_model}
\baike{
Consider $\bar\Delta(t_i) =\log (\bar{I}(t_i))-\log(\bar{I}(t_{i}-\eta))$, and $\bar{I}(t)>0$, for all $(t_i-\eta)\in \mathbb{R}_{>0}$.
If the number of infected cases increases during the time interval $[t_1, t_2]$, then for any time $t_i$ such that $t_i, t_i-\eta \in [t_1, t_2]$, we have $\bar\Delta(t_i) > 0$, and vice versa. If the number of infected cases remains unchanged, then $\bar\Delta(t_i) = 0$.}    
\end{proposition}
\begin{remark}
\label{remark_Delta_function}
\final{The proof of Proposition~\ref{Prop_model} is in the Appendix~\cite{she2023modeling}.}
Proposition~\ref{Prop_model} is a direct result of the definition of $\bar\Delta(t_i)$ and the properties of the logarithmic function.
Note that $\bar\Delta(t_i) =\log (\bar{I}(t_i)/\bar{I}(t_{i}-\eta))$ captures the ratio between the number of infected cases at time step $t_i$ and  time step $t_{i}-\eta$, for $\eta>0$.
We use the function $\bar{\Delta}: \mathbb{R}_{\geq 0}\rightarrow\mathbb{R}$ to model and analyze spreading trends. 
Similar to the reproduction number~\cite{van2017reproduction}, which uses the threshold of one to determine whether infected cases are increasing or decreasing, we can use the threshold of zero for $\bar{\Delta}$ to assess epidemic spread. If $\bar{\Delta}$ is greater than zero, the spread is increasing; if it is less than zero, the spread is decreasing.  
\end{remark}
To simplify the analysis, we consider ${I(t)>0}$ for all ${t\geq 0}$.
In addition, we use i.i.d. noise $\varepsilon(t_i)\sim\mathcal{N}(0, \sigma^2)$ to capture the noise 
term in~\eqref{eq:state}. \baike{Thus, the covariance matrix of the noise terms is $\Sigma = \sigma^2 I_n$.}
For an epidemic spreading process, 
we consider a set of $n$ time steps $T=\{t_1, t_2,\dots, t_n\}$ as the input batch, for $n\in\mathbb{N}_{>0}$.
The corresponding output batch of  $n$ entries is given by $\{\Delta(t_1), \Delta(t_2),\dots, \Delta(t_n)\}$.
We define
$\boldsymbol{\Delta}=[\Delta(t_1), \Delta(t_2),\dots, \Delta(t_n)]^\top$, 
\baike{where $\Delta(t_i)$ is defined in~\eqref{eq:state}, for $t_i\in T$ and $\eta>0$.}
Our goal is to model and predict the first-order difference on the  logarithmic scale of the infected cases. Hence, let the testing location
be the time step $t^*\in \mathbb{T}$,
$\mathbb{T}=[a,b]$.
\baike{We can apply Proposition~\ref{Prop_GPR}
to this model and the results are in Proposition~\ref{Prop_Model_Update}.}

\begin{proposition}
\label{Prop_Model_Update}
Consider an unknown function $\bar\Delta: \mathbb{R}_{\geq 0} \rightarrow \mathbb{R}$.
 The posterior mean at time $t^*$ is given by $m_{\Delta}(t^*) = K(t^*,T)(K(T,T)+\sigma^2I_{n})^{-1}\boldsymbol{\Delta}$, and the posterior variance 
 is $\sigma^2_{\Delta}(t^{*}) = k(t^*,t^*) - K(t^*,T)(K(T,T)+\sigma^2I_{n})^{-1}K(t^*,T)^{\top}$, \baike{where $K(T,T)$ is the kernel matrix between the training points,\final{~$k(t^*,T)$ is the vector of covariances between the testing point $t^*$ 
 and all training points,}~$k(t^*,t^*)$ is the variance at the testing point, and $\sigma^2I_n$ is the covariance matrix of the \final{i.i.d} noise.}
\end{proposition}
For analysis, we specify the kernel function as
the squared exponential kernel to capture the covariance between any pair of points 
$a,b\in\mathbb{R}_{\geq 0}$. The kernel itself is
\begin{equation}
\label{eq_Assum_Ker}
   k(a,b) = \alpha^2 \exp\left(-\frac{(a-b)^2}{2\beta^2}\right),
\end{equation}
 where $\beta>0$  is the length scale of the kernel, and $\alpha^2$ is the signal variance. 



\begin{remark}
While our results can extend to other stationary kernels that are Lipschitz continuous, we use the squared exponential kernel due to its effectiveness in modeling epidemics~\cite{abbott2020estimating}. The squared exponential kernel, commonly used in Gaussian process regression, depends on the distance between variables rather than their absolute positions, promoting smoothness and Lipschitz continuity in the modeled functions. These properties are essential for deriving error bounds~\cite{williams2006gaussian}, and we will use them below. 
\end{remark}

The Gaussian process model in Proposition~\ref{Prop_Model_Update} \baike{and the spread modeling approach in~\eqref{eq:state}} together address Problem~\ref{prob:GPR}. We use the following example to illustrate the proposed Gaussian process model for modeling epidemic spread, \final{which partially addresses Problem~\ref{prob:comparions}.}.

\begin{example}
\label{Example_1}
We use Gaussian process regression to model the real-world epidemic spread of COVID-19 in the United Kingdom, using infection data from March $1^{st}$, 2022, to February $28^{th}$, 2023~\cite{owidcoronavirus}. 
Figure~\ref{fig_UK_data} shows the daily number of infected cases per million people in the UK population during this period. This dataset is chosen due to its multiple waves of infection and the variability in data size over time. To better capture the changes in daily infected cases, we preprocess the data by applying a thirty-day rolling average. For example, the average number of cases on March $30^{th}$, 2022, is calculated by averaging the cases from March $1^{st}$ to March $30^{th}$. 

By indexing March $30^{th}$, 2022 as day one, we represent the daily infected cases on day $t$ as $I(t)$, with $t \in \underline{335}$. 
We set $\eta = 7$. Then 
the difference on the  logarithmic scale of consecutive daily observed infected cases is defined as $\Delta(t) = \log (I(t)) - \log(I(t-7))$, $t\geq 8$, $t \in \underline{335}$. For example, $\Delta(8)$ on April $6^{th}$, 2022, is calculated by \baike{subtracting} 
the logarithmic scale of
the
infected cases on April $6^{th}$ \baike{from the logarithmic scale of those of} March $31^{st}$. We plot $\Delta(t)$, for $t\geq 8$ and $t \in \underline{335}$, from April $6^{th}$, 2022 to February $28^{th}$, 2023 in Figure~\ref{fig:GP_Model} (dotted \final{yellow} line). Comparing Figure~\ref{fig:GP_Model} to Figure~\ref{fig_UK_data}, we see that $\Delta(t)$ effectively captures the trend in the spread. For instance, when $\Delta(t)$ is less than zero from April 2022 to May 2022, the trend in the number of infected cases decreases, as shown in Figure~\ref{fig_UK_data}. This example demonstrates that when $\Delta$ is positive, daily infected cases are increasing, and when it is negative, they are decreasing.
\begin{figure*}[!t]
    \centering
\includegraphics[clip, width=1\textwidth]{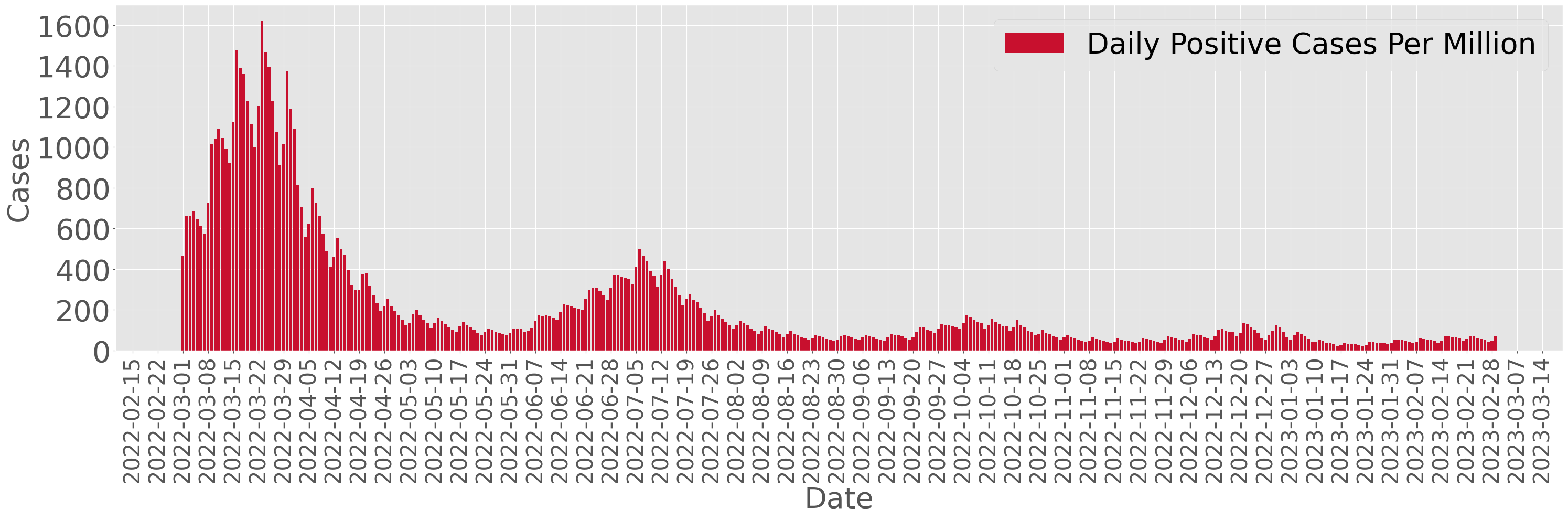}
    \caption{The daily infected cases in the United Kingdom from $03-01-2022$ to $02-28-2023$~\cite{owidcoronavirus}.}
\label{fig_UK_data}
\end{figure*}
\begin{figure*}[!t]
    \centering
\includegraphics[clip,width=1\textwidth]{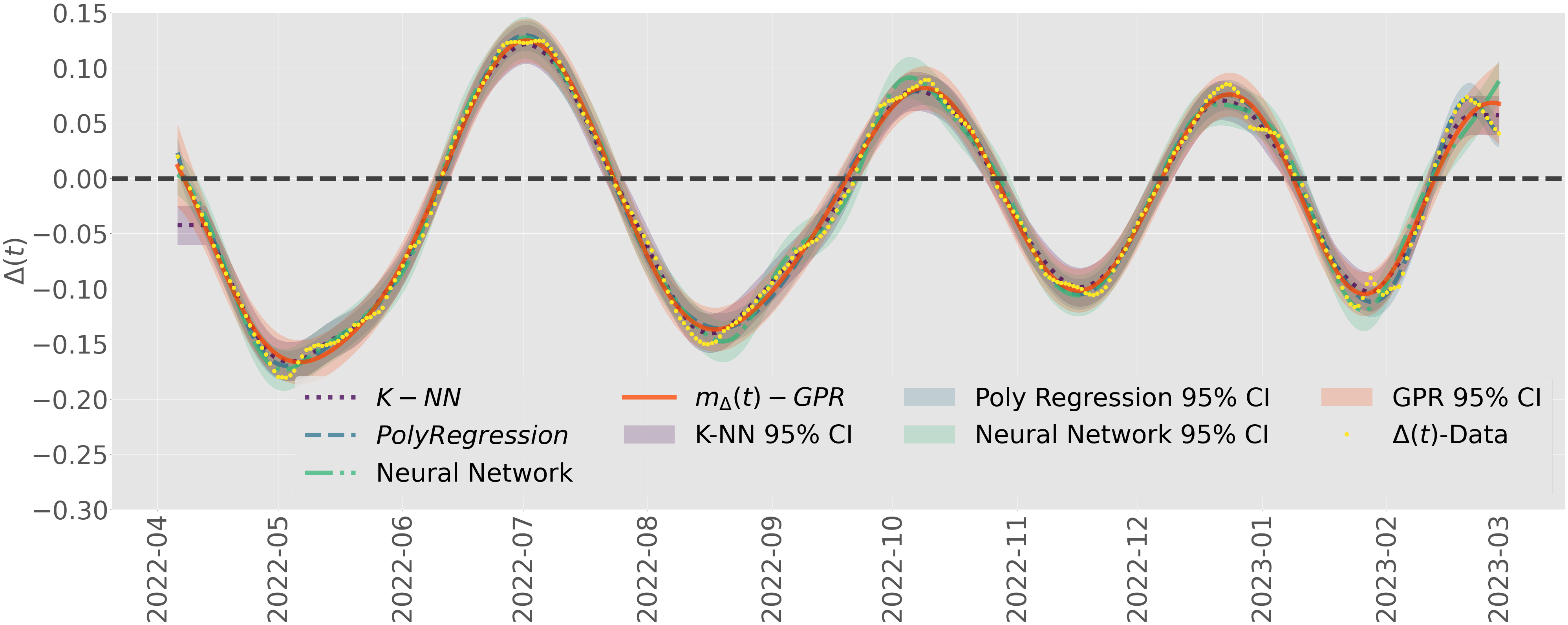}
\caption{Gaussian process regression model for $\Delta(t)$.}
\label{fig:GP_Model}
\end{figure*}

Next, we apply GPR to model the spread. The training \baike{time steps} are $\{8, 9, \dots, 365\}$, corresponding to the training data $\{\Delta(8), \Delta(9), \dots, \Delta(365)\}$, as depicted by the \final{dotted yellow} line in Figure~\ref{fig:GP_Model}.  Using the GPR algorithm from Proposition~\ref{Prop_Model_Update}, we visualize the posterior means at the training time steps in Figure~\ref{fig:GP_Model}.
We use $\varepsilon \sim \mathcal{N}(0, 0.002)$ \baike{to ensure that $98.18\%$ of the noisy training data points fall within the $95\%$-confidence interval.}
The \final{red solid line} 
in Figure~\ref{fig:GP_Model}
represents the posterior mean, denoted as $m_{\Delta}^*(t)$, at the training locations $t \in \underline{365}$ and $t\geq 8$, while the shaded \final{red} region shows the $95$\%-confidence interval. \final{We observe that $98.18\%$ of the training data are within the $95\%$ confidence intervals.}
\final{We compute the mean square error (MSE) using the posterior mean and the training data, resulting in an MSE of $0.000069$ for the Gaussian process regression model.}
The example shows how modeling the difference on the  logarithmic scale of the infected cases effectively captures the spread dynamics and highlights the utility of Gaussian process regression in doing so.

To validate the GPR model, we first leverage polynomial regression and $k$-nearest neighbors (KNN) to model the same data, as these methods are commonly used as comparisons to disease spreading modeling methods~\cite{senanayake2016predicting}. We select the model parameters for polynomial regression and KNN to achieve relatively accurate models, based on the MSE and the percentage of training data that fall within the $95\%$ confidence intervals. 
For polynomial regression, we use a degree of $20$, resulting in an MSE of $0.000048$, with $93.64\%$ of the training data within the $95\%$ confidence intervals. The model generated by polynomial regression is represented by a dashed blue line, with the $95\%$ confidence interval captured by blue shaded areas.
For the KNN model, we set 
$\kappa=15$, resulting in an MSE of $0.000080$, with $95.78\%$ of the training data within the $95\%$ confidence intervals. The model generated by KNN is represented by a dotted purple line, with the $95\%$ confidence interval captured by purple shaded areas.

In addition, we leverage a neural network to model the same data. The data exhibits periodic patterns due to the fluctuations in the logarithmic scale of the infected cases. Given that standard neural networks, without proper feature augmentation, may struggle to efficiently learn such periodic behavior, we first apply sinusoidal feature generation. This technique augments the input data by incorporating sine transformations of the features~\cite{wong2022learning}.
We use a three-layer neural network (with $200$ neurons in the input layer, $150$ neurons in the hidden layer, and $100$ neurons in the output layer). The MSE for the neural network is $0.000090$, with $93.33\%$ of the training data falling within the $95\%$ confidence intervals. The model generated by this neural network is represented by a dash-dotted green line, with the $95\%$ confidence interval captured by green shaded areas.

All four modeling approaches can generate relatively accurate models, with MSEs within the same order of magnitude. Additionally, most of the training data fall within the $95\%$ confidence interval. However, compared to GPR, which requires selecting an appropriate kernel function, the other three methods require parameter tuning: the polynomial order for polynomial regression,  
$\kappa$ for KNN, and the number of layers and neurons for neural networks.
Moreover, while GPR allows flexibility in kernel selection, such as replacing the squared exponential kernel with periodic kernels when periodic patterns are observed~\cite{senanayake2016predicting}, this is not strictly necessary in modeling. Without sinusoidal feature generation for the input data, the neural network model with the current configuration of layers and neurons would underfit, resulting in an MSE three orders of magnitude higher than the current results. Furthermore, although all three alternative models can capture the spread, their more complex structures may lead to overfitting, as will be discussed in Example~\ref{Example_3}.
\end{example}
\subsection{The Impact of Spread Data on Prediction Variance}
\label{Sec_Epi_Prediction_Var}
We next propose an upper bound for the prediction variance to study the impact of data on predictions.  
\begin{lemma}
\label{Lem_var}
Consider a set of time steps $T=\{t_1, t_2,\dots,t_n\}$. 
\baike{Let $\mathbb T_r(t^*)$
 represent all points in $T$ that lie within a ball of radius $r$, centered at the testing location $t^*$, i.e., $\baike{\mathbb{T}_r(t^*) = \{t \in T : |t - t^*| \leq r \}.}$
}
    Consider the squared exponential
    kernel from~\eqref{eq_Assum_Ker}
    for the Gaussian Process model in Proposition~\ref{Prop_Model_Update}.
    The posterior variance $\sigma^2_{\Delta}(t^{*})$ at the testing time step $t^*$ obeys the bound
    $\sigma^2_{\Delta}(t^*) \leq \alpha^2 - \frac{\alpha^2}{\alpha^2+\frac{\sigma^2}{|\mathbb T_{r}(t^*)|}}$.
\end{lemma}

\final{The proof of Lemma~\ref{Lem_var} is provided in the Appendix~\cite{she2023modeling}.}
Lemma~\ref{Lem_var} provides a data-dependent posterior variance bound on $\sigma^2_{\Delta}(t^*)$ at the testing time step $t^*$ in terms of the number of the training data points around the testing location. 
The bound on the prediction variance given by Lemma~\ref{Lem_var} shows that more training data in $\mathbb{T}_r(t^*)$, captured by $|\mathbb{T}_r(t^*)|$, leads to a lower variance bound. 
Additionally, higher 
variance noise in the data, captured by~$\sigma^2$, 
increases the bound
on the posterior variance. 
In the absence of observation noise or with infinitely many data points near $t^*$, the variance bound approaches zero. 
Lemma~\ref{Lem_var} illustrates how the available data, the 
spreading trend, and noise can affect the posterior variance bound, and it solves Problem~\ref{prob:data}.
\subsection{High Probability Error Bound on the Prediction}
\label{Sec_Epi_Prediction_Error}
\bshe{Having discussed the impact of data on the upper bound of the posterior variance, we now analyze the error bound on the prediction error.}
We first introduce the Lipschitz constant of the squared exponential kernel~\cite{lederer2021uniform}. We consider the space of sample functions corresponding to the space of continuous functions on the time interval $\mathbb T\subset [t_a, t_b]$, where $t_a,t_b\in\mathbb{R}_{>0}$, such that the input batch $T\subset \mathbb T$. 
\begin{lemma}\cite[Corollary 8]{lederer2021uniform}
\label{Lem_Lip}
\baike{Consider $t_i, t_j\in \mathbb T$. 
Then for all~$t$, 
$k(\cdot, t)$ is~$L_k$-Lipschitz, 
where~$L_k = \frac{\alpha^2}{\beta e^{1/2}}$.} 
\end{lemma}

We see that $L_k$ is determined by the kernel parameters $\alpha$ and $\beta$ in~\eqref{eq_Assum_Ker}. 
In addition, consider our continuous unknown function 
\baike{$\bar{\Delta}$ from Remark~\ref{remark_Delta_function}, where}
$\bar{\Delta}:\mathbb{T}_{\geq0} \rightarrow\mathbb{R}$
with Lipschitz constant
$L_{\bar\Delta}$, such that $|\bar{\Delta}(t_1)-\bar{\Delta}(t_2)| \leq L_{\bar\Delta} \left|t_1-t_2\right|$ for all $t_1, t_2\in \mathbb T$. The Lipschitz continuity of
$\bar{\Delta}(t)$ indicates that the change in spread over some time interval is limited by the
length of that interval.

Recall the training dataset $\{(t_i,\Delta(t_i))\}_{i=1}^n$ where  $t_i\in T$.
We consider the training dataset to be within a specific time interval of interest, i.e., $T\subset \mathbb{T}$, and
the testing point
is also in this interval, i.e., $t^*\in\mathbb{T}$. 
The time interval of interest may be extensive, causing the testing location $t^*$ to be far from any of the training points in the dataset $T$. 

\begin{figure*}
    \centering  \includegraphics[clip, width=1\linewidth]{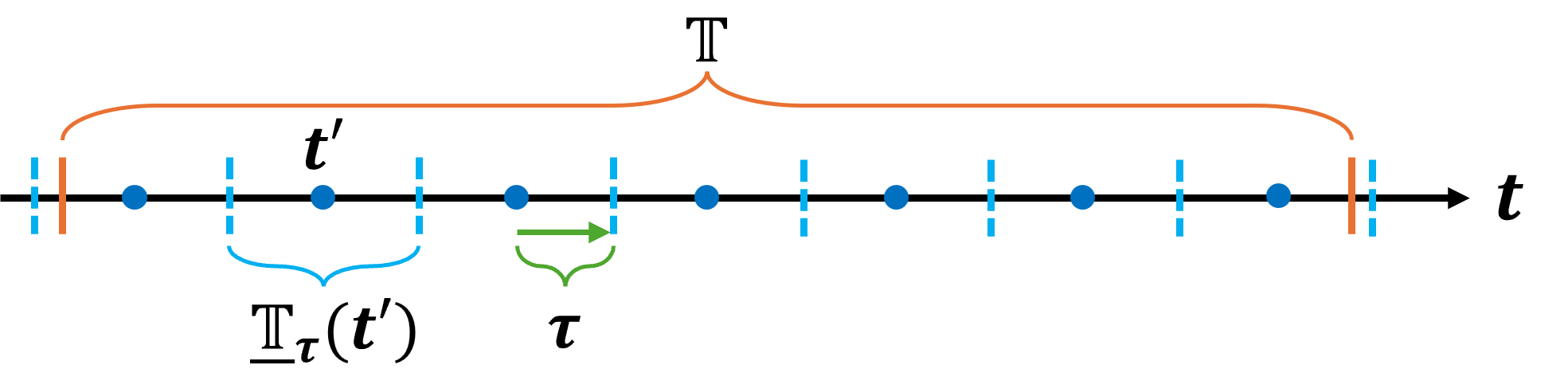}
    \caption{A time interval $\mathbb{T}$ with  $M(\tau,\mathbb{T})=7$.}
    \label{fig:covering_number}
\end{figure*}

To address this issue when studying the posterior mean, we
consider a set of grid points $\mathbb{M}$
that are evenly distributed on $\mathbb{R}_{\geq 0}$. 
\baike{We define the radius associated with the point $t'\in\mathbb{M}$
as $\tau$, such that $\underline{\mathbb{T}}_{\tau}(t'):=[t'-\tau,t'+\tau]$ for all $t'\in \mathbb{M}$.}
Then
we consider $T\subset\mathbb T\subseteq\{t \mid t\in \underset{ t'\in \mathbb{M}}{\bigcup}[t'-\tau,t'+\tau]\}$. 
To ensure that the union of the intervals of length~$2\tau$ can cover the continuous-time interval $\mathbb T$, 
the length of the interval 
between any two neighboring points in $\mathbb{M}$
must be smaller than  $2\tau$. 
We define the covering number of $\mathbb T$, denoted by 
$M(\tau, \mathbb T)$, as the
cardinality of the set $\mathbb{M}$ with the minimum number of grid points to satisfy $\mathbb{T}\subseteq\{t \mid t\in \underset{ t'\in  \mathbb{M}}{\bigcup}[t'-\tau,t'+\tau]\}$. 
Figure~\ref{fig:covering_number} illustrates a time interval $\mathbb{T}$ with $M(\tau, \mathbb T) = 7$.

\baike{
We derive the following theorem 
in part by using results from~\cite{lederer2019uniform}. 
This result is gives a high-probability bound on 
prediction accuracy, and it solves 
 Problem~\ref{prob:error}.}
\begin{theorem}
\label{Them_Uniform_Bound}
Consider a zero-mean Gaussian process defined through the kernel $k(\cdot,\cdot)$ in~\eqref{eq_Assum_Ker} with Lipschitz constant $L_k$ on time interval  $\mathbb{T}$.
Consider a
continuous unknown function $\bar\Delta: \mathbb{T}_{\geq 0}\rightarrow \mathbb{R}$ that  that captures the spreading process through \final{the difference on the logarithmic scale of the infected cases}, 
with Lipschitz constant $L_{\bar\Delta}$.
The posterior mean  of a Gaussian process conditioned on the training dataset
$\{(t_i,\Delta(t_i))\}_{i=1}^n$, for all
$t_i\in T$, at the testing time step $t^*$ is given by $m_{\Delta}(t^*)$. 
Consider a set of grid points $\mathbb{M}$. 
If, for all $t^*\in\mathbb{T}$, there exists  $t_i\in T$ and $t'\in \mathbb{M}$ such that $t_i\in\underline{\mathbb{T}}_{\tau}(t')$ and $|t^*-t_i|\leq\tau$, then
\begin{align*}
 \nonumber
 & P\left(\left| \bar{\Delta}(t^*)-m_{\Delta}(t^*)\right|
\leq \sqrt{\gamma_{\delta}(\tau)}\sigma_{\Delta}(t^*)+\xi_{\delta}(\tau)\right)\geq 1-\delta, 
\end{align*}
for all $\delta\in(0,1)$.
Note that $\bar{\Delta}(t^*)$ is the noise-free variable at the testing time step $t^*$,
and $m_{\Delta}(t^*)$ and $\sigma_{\Delta}(t^*)$ are the posterior mean and posterior standard deviation from Proposition~\ref{Prop_Model_Update}, respectively. 
Further, we have 
\begin{align}
\gamma_{\delta}(\tau) &=
2\log\left(\frac{\hat{\mathbb T}}{2\tau\delta}+\frac{1}{\delta}\right) \\
\xi_{\delta}(\tau) &= (L_{\bar\Delta} +L_m)\tau + \sqrt{\gamma_{\delta}(\tau)L_{\sigma^2}\tau},
\end{align}
where
$L_{\bar\Delta}$ is the Lipschitz constant of the function $\bar{\Delta}$, 
$L_m=\frac{\alpha^2}{\beta e^{1/2}}\sqrt{n}\|(K(T,T)+\sigma^2I_n)^{-1}\boldsymbol{\Delta}\|$ is the Lipschitz constant of the posterior mean function \bshe{$m_{\Delta}$}, 
and 
$L_{\sigma^2} =  \frac{2n\alpha^4}{\beta e^{1/2}} \left\|(K(T,T)+\sigma^2I_n)^{-1}\right\|$
is the Lipschitz constant of the posterior
variance function $\sigma^2_{\Delta}$.
\end{theorem}

\final{The proof of Theorem~\ref{Them_Uniform_Bound} is provided in the Appendix~\cite{she2023modeling}.} This theorem gives a high-probability error bound on the prediction error, which depends on the distance between training and testing points, captured by the length of the time interval $\hat{\mathbb{T}}$. A larger $\hat{\mathbb{T}}$ results in a larger error bound, and vice versa. 
The bound also shows that higher posterior variance at the testing time step $\sigma_{\Delta}(t^*)$
increases the error bound. 
Additionally, a lower Lipschitz constant of the spreading function $L_{\bar\Delta(t)}$ reduces the error bound, as smaller changes in the spreading process correspond to a lower Lipschitz constant. While Proposition~\ref{Prop_Model_Update} provides an analytical solution for the posterior mean and variance of the GPR model, the bounds from Lemma~\ref{Lem_var} and Theorem~\ref{Them_Uniform_Bound} offer valuable insights for selecting datasets and data collection methods, helping to improve modeling outcomes.

\begin{figure*}[!t]
    \centering
\includegraphics[clip, width=1\linewidth]{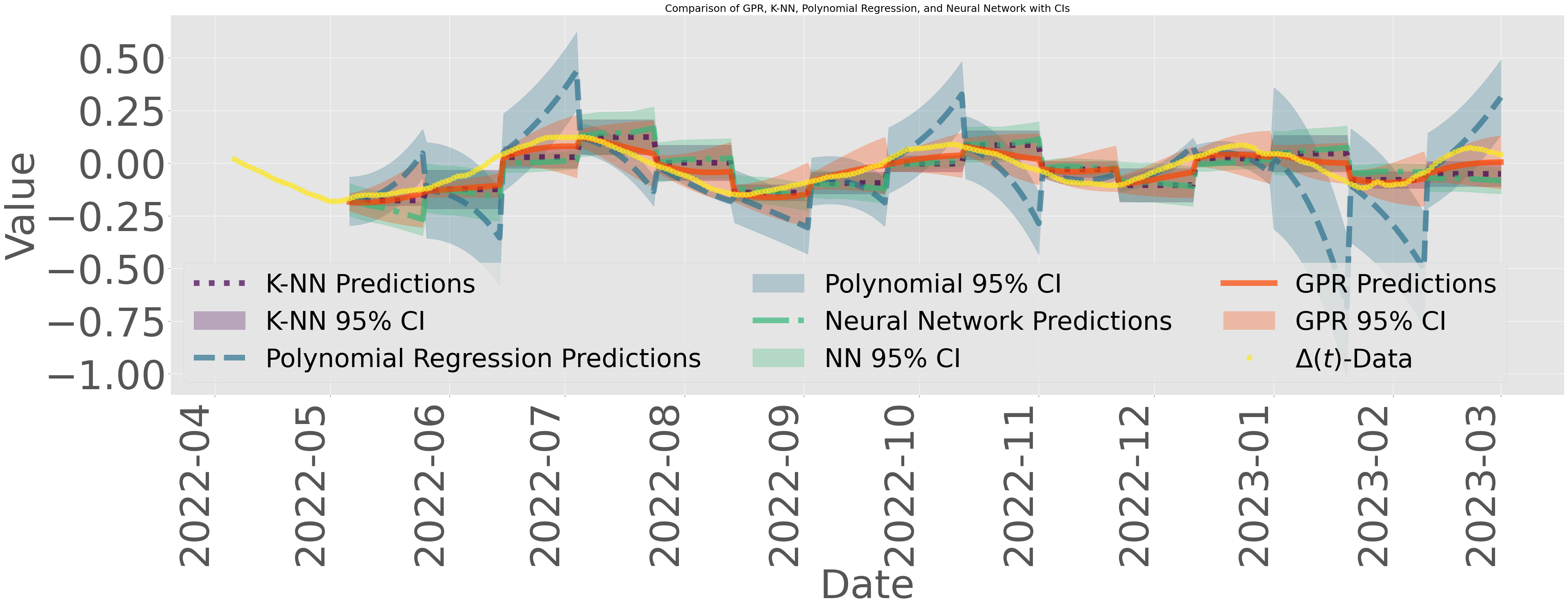}
\caption{Posterior mean of the prediction on spreading trend. Since we lack additional information when modeling the spread, such as periodic spreading patterns, we observe lower prediction accuracy when $\Delta$ shifts from increasing to decreasing, or vice versa.} 
\label{fig:GPR_Pred}

\end{figure*}

\begin{figure*}[!t]
    \centering
\includegraphics[clip, width=1\linewidth]{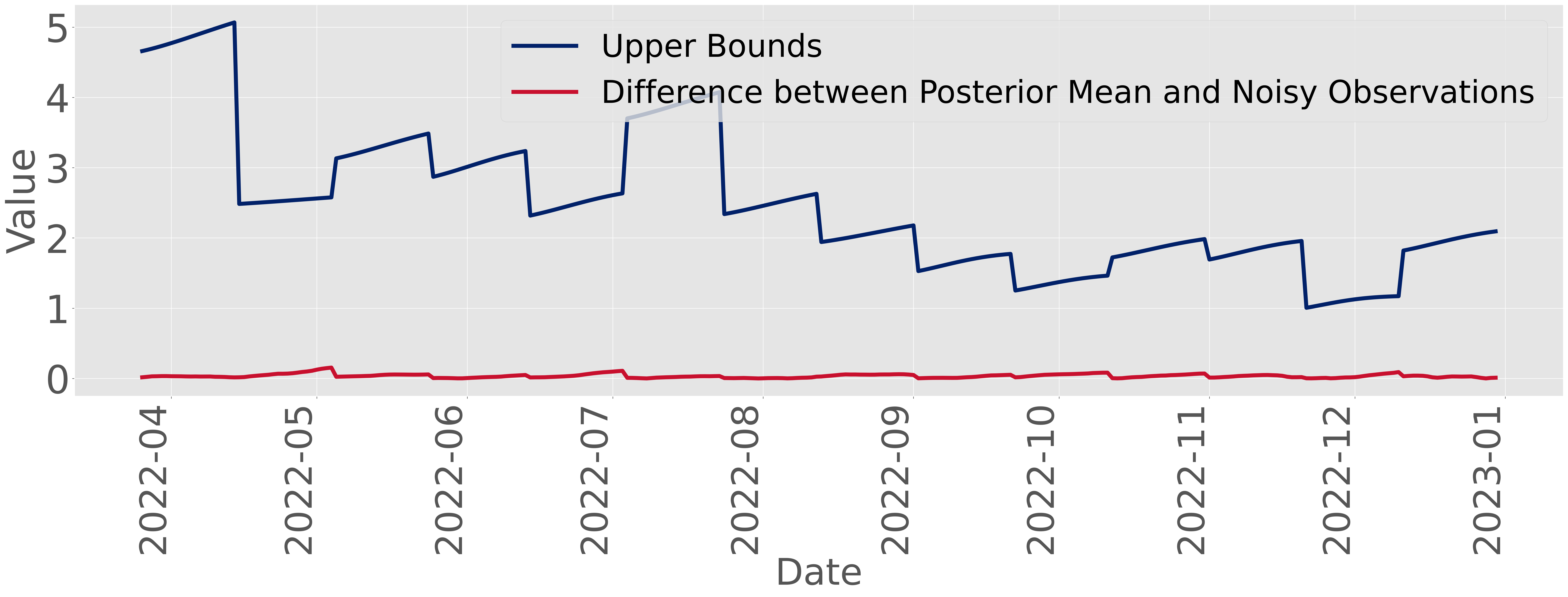}
\caption{Probability upper bound with $\delta = 0.05$. The upper bound changes depending on how far into the future the prediction length extends. For the same twenty-day period, the further into the future, the larger the prediction error bound will be.}
    \label{fig:GPR_Pred_High_Bound}
\end{figure*}

\begin{example}
\label{Example_3}
Consider the dataset representing the difference on the  logarithmic scale of the infected cases in the UK, shown by the solid red line in Figure~\ref{fig:GP_Model}. 
\final{As in Example~\ref{Example_1}, the dataset is preprocessed using a thirty-day rolling average, with $\eta$ set to 7.}
Then,
we use a moving window of $30$ days of data to train the GPR model and predict the posterior mean for the subsequent $20$ days. For example, data from April $6^{th}$ to May $6^{th}$, 2022, is used to predict from May $7^{th}$ to May $27^{th}$, 2022.
In Figure~\ref{fig:GPR_Pred}, the  solid \final{red} line represents the posterior mean of the predictions, with the shaded \final{red} area showing the $95\%$ confidence interval. The \final{dotted yellow} line represents the noisy data. 
\baike{Around $94.29\%$ of noisy data are located within the $95\%$ confidence interval. Most predictions that fail to include the data within the 95\% confidence interval
arise
from either abrupt changes in the spreading trend and/or the
limited number of available data samples.}
This figure illustrates that the prediction captures the trend of the spread when compared to the noisy observations. Note that the confidence interval represents the uncertainty in predicting the true value, which is not directly available.


\final{We further compare the GPR prediction results with the modeling methods considered in Example~\ref{Example_1}, including polynomial regression, KNN, and neural networks.
First, we use polynomial regression of degree three, resulting in an MSE of $0.0183$, with $72.00\%$ of the testing data falling within the $95.00\%$ confidence intervals. The predictions generated by polynomial regression are represented by a dashed blue line, with the $95\%$ confidence interval captured by blue shaded areas.
For the KNN model, we set $\kappa = 3$, which yields an MSE of $0.0046$, with $51.00\%$ of the testing data falling within the $95\%$ confidence intervals. The KNN model predictions are represented by a dotted purple line, with the $95\%$ confidence interval captured by purple shaded areas.}

\final{In addition, we leverage a standard neural network to predict the same spread, where sinusoidal feature generation is not applied to the input, as we only use thirty data points to train the model for each time period.
The neural network consists of three layers: an input layer with $50$ neurons, a hidden layer with $25$ neurons, and an output layer with $10$ neurons. The MSE for predictions generated by the neural network is $0.0070$, with $43.67\%$ of the testing data falling within the $95\%$ confidence intervals. The model generated by this neural network is represented by a dash-dotted green line, with the $95\%$ confidence interval captured by green shaded areas.}

\final{We further discuss why different models are used for prediction compared to the modeling process in Example~\ref{Example_1} for polynomial regression, KNN, and neural networks. In the modeling process, capturing the periodic pattern of the spread over an entire year requires higher orders for both polynomial regression and KNN. However, using higher orders for these models can lead to overfitting in the prediction task.
Specifically, when leveraging only thirty days of data to predict the spread over the next twenty days, setting the polynomial and KNN orders to $20$ and $15$, respectively, results in severe overfitting and high MSE. Consequently, we adjust the orders for both models to improve their performance in the prediction task.}
\final{For prediction using neural networks, instead of employing the same model with three layers of neurons (an input layer with $200$ neurons, a hidden layer with $150$ neurons, and an output layer with $100$ neurons) and sine transformations of the training data features, we use a standard neural network with three layers of neurons (containing $50$, $25$, and $10$ neurons in the input, hidden, and output layers, respectively).
We avoid feature augmentation and reduce the number of neurons in each layer to prevent overfitting, given the significantly smaller amount of data and the absence of periodic patterns within each training and prediction period.}

\final{Even after adjusting the parameters for polynomial regression, KNN, and neural networks, resulting in MSEs of the same order of magnitude as the GPR, the proportion of predictions within the $95\%$ confidence interval is significantly lower than that of the GPR predictions.
More importantly, the same GPR modeling approach used in Example~1 
can effectively capture the periodic patterns in the spread. In 
contrast, the other three methods require parameter adjustments to 
balance modeling and prediction performance, making them less 
robust for capturing such complex dynamics.}

In addition to comparing to other prediction models, Figure~\ref{fig:GPR_Pred_High_Bound} illustrates the difference between the posterior mean and the noisy observations at each prediction, shown by the  solid red line. The upper bound, computed based on Theorem~\ref{Them_Uniform_Bound}, is plotted in blue. We assume that
$M(\mathbb{T},\tau) = 5$, where $\tau = 5$ and  $\hat{\mathbb{T}}=50$. We assume that
the Lipschitz constant of the function $\Delta$ is $L_{\bar\Delta} = 0.01$ and $\delta=0.05$. 
We observe that, during each twenty-day prediction period, the bound increases monotonically as the prediction time steps move further away from the training data. By comparing the bound from April 2022 to September 2022 with the bound from September 2022 to March 2023, we find that the bound is looser during periods of significant changes in the spread (April 2022 to September 2022), leading to a more conservative bound. Conversely, when the spread is more consistent (September 2022 to March 2023), the bound tightens. The bound is generally larger when considering a longer time interval into the future. However, the properties of the bound can  guide us in selecting appropriate datasets and data collection methods to improve prediction results.
\end{example}
\section{Sensitivity Analysis}
\label{Sec_Sen_analysis}
This section serves as supporting information for performing a sensitivity analysis to highlight the potential impact of data preprocessing on the empirical results. We include two discussions of the sensitivity analysis, regarding the GPR modeling in Example~\ref{Example_1} and GPR prediction in Example~\ref{Example_3}. Detailed information can be found at the link: \url{https://github.com/baikeshe/GPR_Epi_Modeling.git}.

\subsection{Sensitivity Analysis on GPR Modeling}
\begin{figure*}
    \centering
\includegraphics[trim = 0cm 0cm 0cm 0cm, clip, width=1\textwidth]{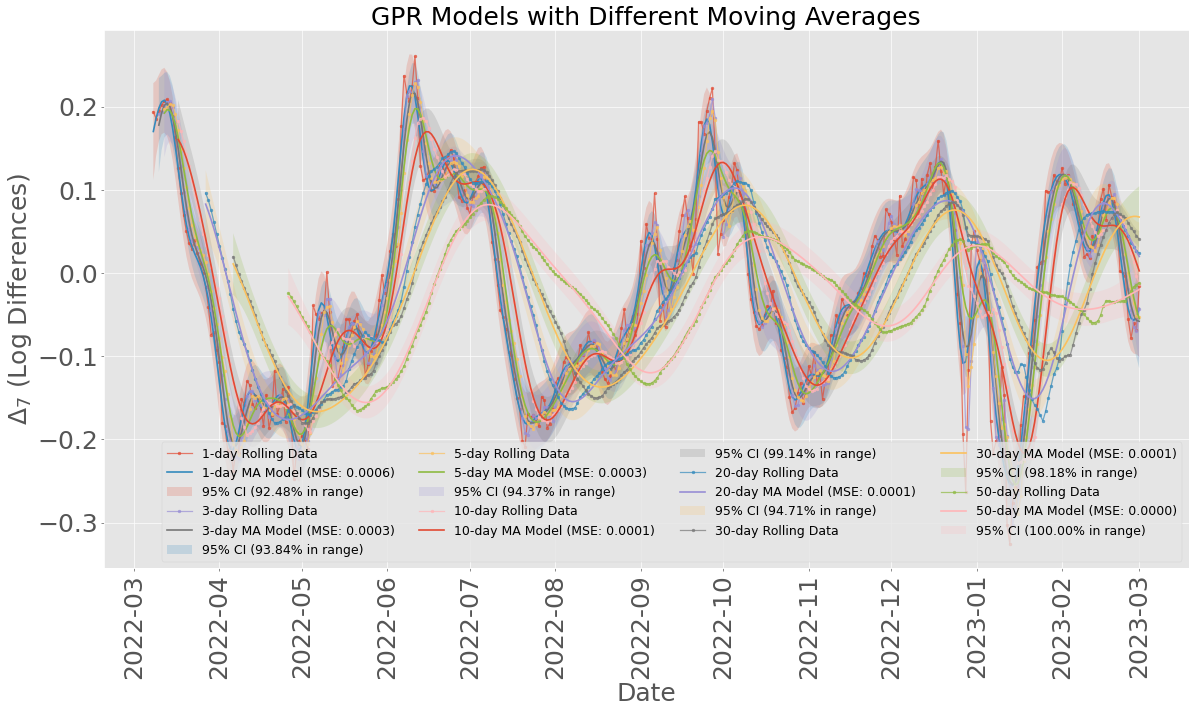}
    \caption{GPR model with $1$-, $3$-, $5$-, $10$-, $20$-, $30$-, and $50$-day smoothing window and $\eta=7$. Shorter smoothing windows capture short-term perturbations, while longer smoothing windows reveal more pronounced periodic patterns. Despite these differences, all GPR models retain at least $92.48\%$ of the training data within the $95\%$ confidence intervals.}
\label{fig_GPR_Model_Varying_Size}
\end{figure*}

\begin{figure*}
    \centering
\includegraphics[trim = 0cm 0cm 0cm 0cm, clip,width=1\textwidth]{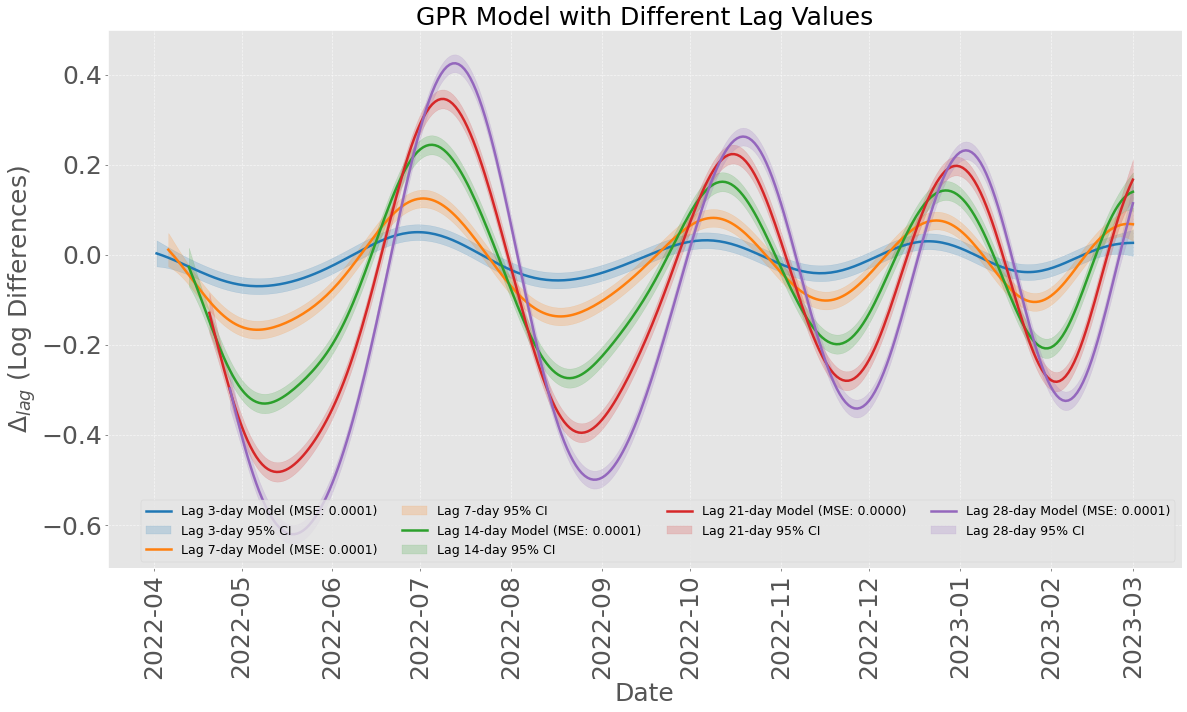}
\caption{GPR model with $30$-day smoothing window and $\eta$ is selected from $\{1, 3, 7, 14, 21, 28\}$. Higher values of 
$\eta$ result in narrower prediction confidence intervals because larger 
$\eta$ better captures longer-term spreading trends, thereby reducing noise over shorter time periods.}
\label{fig_GPR_Model_Varying_Lag}
\end{figure*}

Instead of leveraging a fixed window of $30$, we conduct a sensitivity analysis using different lengths for the averaging window in GPR modeling. 
First, we perform sensitivity analyses for data preprocessing in modeling the spread using GPR for the simulation in Example~\ref{Example_1}. In addition to using a $30$-day smoothing window and a $7$-day difference on the logarithmic scale of infected cases, we conduct two types of sensitivity analysis for the modeling. In the first scenario, 
we fix the $7$-day difference on the logarithmic scale and varied the $30$-day smoothing window to $1$, $3$, $5$, $10$, $20$, $30$, and $50$ days, using the same GPR method to model the data. The results are illustrated in Figure~\ref{fig_GPR_Model_Varying_Size}, where shorter smoothing windows capture short-term perturbations, while longer smoothing windows reveal more pronounced periodic patterns.  Despite these differences, all GPR models retain at least $92.48\%$ of the training data within the $95\%$ confidence intervals.
In the second scenario, we fix the $30$-day smoothing window and vary the $7$-day difference to $1$, $3$, $7$, $14$, $21$, and $28$ days, comparing the modeling results using the same GPR approach. The results are shown in Figure~\ref{fig_GPR_Model_Varying_Lag}, where higher values of 
$\eta$ result in narrower prediction confidence intervals because larger 
$\eta$ better captures longer-term spreading trends, thereby reducing noise over shorter time periods. 

\begin{figure*}[!h] 
    \centering
\includegraphics[trim = 0cm 0cm 0cm 0cm, clip, width=1\textwidth]{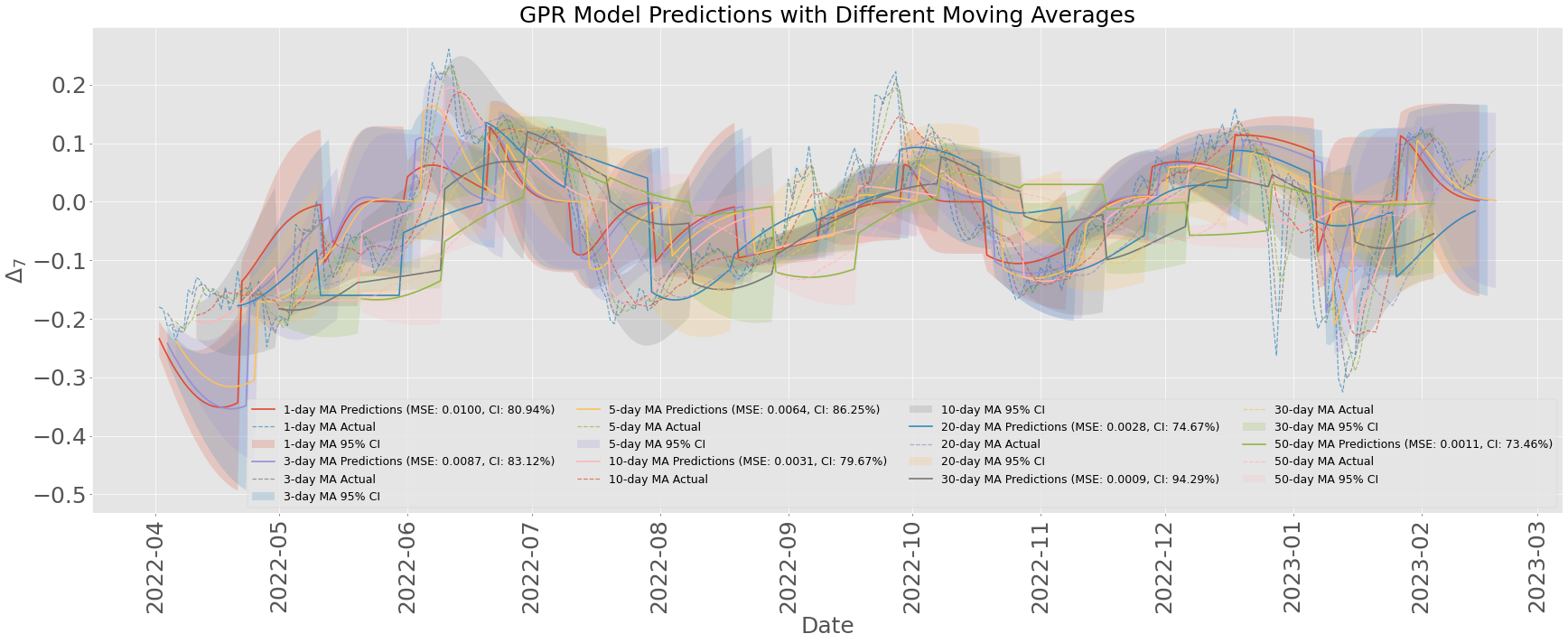}
    \caption{GPR Prediction with $1$-, $3$-, $5$-, $10$-, $20$-, $30$-, and $50$-day smoothing window and $\eta=7$. A $30$-day moving average of the data provides an optimal balance for data preprocessing, capturing periodic trends while filtering short-term noise. Shorter windows ($1,3,5$ days) retain detail but risk overfitting, while moderate windows ($10, 20$ days) struggle to capture both short- and long-term dynamics. Overly long windows ($50$ days) oversmooth the data, reducing model adaptability.}
\label{fig_GPR_Pre_Varying_Size}
\end{figure*}
\begin{figure*}[!h]
    \centering
\includegraphics[trim = 0cm 0cm 0cm 0cm, clip,width=1\textwidth]{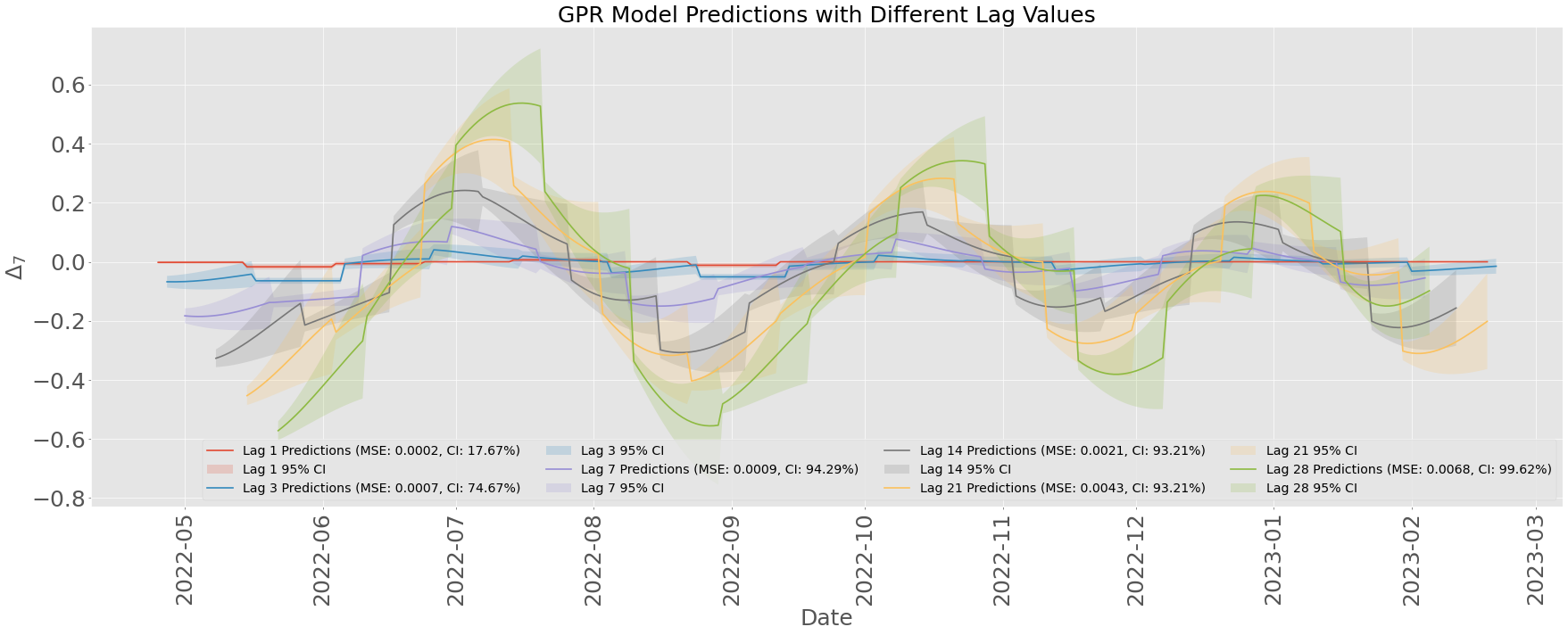}
\caption{GPR Prediction with $30$-day smoothing window and $\eta$ is selected from $\{1, 3, 7, 14, 21, 28\}$. GPR Prediction with a fixed $30$-day smoothing window while varying the length of the log-difference on infected cases, where $\eta$ was selected from the set $\{1, 3,7,14,21,28\}$.}
\label{fig_GPR_Pre_Varying_Lag}
\end{figure*}

\subsection{Sensitivity Analysis on GPR Prediction}
After discussing the impact of the data preprocessing on 
GPR modeling, we perform sensitivity analyses for data preprocessing in
predicting the spread using GPR for the simulation in Example~\ref{Example_3}. 
Instead of leveraging a fixed window of $30$, we first conduct a sensitivity analysis by varying the averaging window lengths for data-preprocessing for GPR training and prediction. The smoothing window for data preprocessing is draw from the set $\{1,3,5,10,20,30,50\}$, where we fixed $\eta = 7$. The GPR prediction results are given in Figure~\ref{fig_GPR_Pre_Varying_Size}. Figure~\ref{fig_GPR_Pre_Varying_Size} shows that when the window sizes are $1$, $3$, and $5$, the prediction accuracies are $80.94\%$, $83.12\%$, and $86.25\%$, with corresponding MSE values of $0.0100$, $0.0087$, and $0.0064$, respectively. 
We also experiment with smoothing window sizes of $10$, $20$, $30$, and $50$. The prediction accuracies for these windows are $79.67\%$, $74.67\%$, $94.29\%$, and $73.46\%$ with corresponding MSE values of $0.0031$, $0.0028$, $0.0011$ and $0.0009$, respectively. According to the simulations in Figure~\ref{fig_GPR_Pre_Varying_Size}, we provide the following summary of the prediction results. 
\begin{itemize}
    \item Short Smoothing Windows ($1$, $3$, $5$ days): Shorter smoothing windows retain much of the data's short-term variability. This allows the GPR model to capture and adapt to rapid changes or fluctuations in the time-series data, leading to improved short-term predictions. However, if the window is too short, the model may overfit to noise rather than the true underlying trend, limiting its ability to generalize to unseen data. This is why even smaller windows (like $1$ day) may still result in suboptimal prediction.
    \item Moderate Smoothing Windows ($10$, $20$ days): These window sizes may filter out too much short-term variability while still not fully capturing the longer-term periodic patterns. The model can struggle to find a balance, missing some important details necessary for accurate predictions. The smoothed data using these window sizes might show incomplete or mixed trends, which can confuse the GPR, resulting in less reliable predictions and result in lower prediction accuracy.
    \item Optimal Smoothing Window ($30$ days): A 30-day smoothing window provides the right balance between filtering out short-term noise and capturing long-term patterns, including periodic trends in disease spread. This leads to more stable and accurate predictions. With a 30-day window, the underlying periodic behavior (such as monthly trends in the spread of disease) becomes clearer. This helps the GPR model learn and extrapolate these patterns effectively, enhancing prediction accuracy.
    \item Long Smoothing Windows ($50$ days): A longer smoothing window excessively suppresses short-term variations, overemphasizing long-term trends. This reduces the model's flexibility and sensitivity to recent changes in the data, causing poorer predictions, particularly when abrupt changes occur.
\end{itemize}
Therefore, using a $30$-day moving average for data preprocessing provides an optimal balance. It effectively captures relevant periodic trends while filtering out short-term noise. Shorter windows ($1, 3, 5$ days) may improve predictions by retaining more detail but risk overfitting to noise. In contrast, moderate windows ($10, 20$ days) fail to capture both short-term and long-term dynamics effectively, while overly long windows ($50$ days) oversmooth the data, limiting the model's adaptability. In addition, we provide GPR prediction results with a fixed $30$-day smoothing window while varying the length of the log-difference on infected cases, with $\eta$ selected from the set ${3,7,14,21,28}$. Figure~\ref{fig_GPR_Pre_Varying_Lag} also shows that by selecting an appropriate $\eta$, we can balance prediction accuracy and the spreading patterns observed in the data.
\section{Conclusion}
\label{sec_conclusion}
In this work, we propose an approach to modeling epidemic spreading processes by using Gaussian process regression. We model and predict the spread through the difference on the  logarithmic scale of the infected cases. We provide an upper bound on the posterior variance and mean error, highlighting the impact of the spreading trend and available data. \final{Additionally, we discuss the impact of data preprocessing on modeling and prediction performance. We further highlight the benefits of using GPR methods by empirically comparing them with other modeling and prediction approaches, including polynomial regression, KNN, and neural networks.}
In future work, we plan to utilize this model and prediction mechanism to design a new data-driven predictive control strategy for epidemic mitigation.


\bibliographystyle{IEEEtran}
\bibliography{root}
\newpage
\section*{Appendix}
\subsection{Proof of Proposition~\ref{Prop_model}}
If the number of infected cases is increasing in the time interval $[t_1, t_2]$, and $t_i, t_i-\eta \in [t_1, t_2]$, we have that $\frac{\bar{I}(t_i)}{\bar{I}(t_{i}-\eta)}>1$, where $I(t)>0$ for all $t\in[t_1, t_2]$. According to~\eqref{eq:state}, 
    $\bar\Delta(t_i) =\log (\bar{I}(t_i)/\bar{I}(t_{i}-\eta))
    \geq \log (1) =0$,
where we use $\log (1) =0$ and that the natural logarithm function is monotonically increasing. We can apply the same approach to demonstrate scenarios where the number of infected cases is either decreasing or remains constant over the interval $[t_1, t_2]$.
\qed

\subsection{Proof of Lemma~\ref{Lem_var}}
We first introduce an established result. 
We consider a Gaussian process model with an isotropic kernel $k(\cdot,\cdot)$, which decreases as the distance between its arguments decreases. We
also have 
an input
training dataset $X$ with zero-mean observation noise and a specified variance $\sigma^2$. 
Let \baike{$\mathbb{B}_{r}(x^*)=\{x\in X: \|x-x^*\|\leq r \}$}
denote the training dataset restricted to a ball centered at the testing location $x^*$,  where $x,x^*\in\mathbb{R}^n$. We have that the radius of the ball $r\leq \frac{k(x^*,x^*)}{L_k}$, where $L_k$ is the Lipschitz constant of the kernel function $k(\cdot, \cdot)$ in its first argument.
According to~\cite[Corollary 2]{lederer2021uniform}, the posterior variance of the prediction is bounded by 
$\sigma^2(x^*)\leq k(0,0) - \frac{k^2(r,r)}{k(0,0)+\frac{\sigma^2}{|\mathbb{B}_{r}(x)|}}.$

Now consider the input batch $T=\{t_1,t_2,\dots, t_n\}$ and the output batch  $\{\Delta(t_1), \Delta(t_2), \dots, \Delta(t_n)\}$.
Under the condition that \baike{$\mathbb{T}_r(t^*) = \{t \in T :  
|t - t^*| \leq r
\}$}
contains 
all time steps in~$T$
that are within distance~$r$ of the testing location $t^*$, we have that $\sigma^2_{\Delta}(t^{*})
\leq\alpha^2 - \frac{\alpha^4}{\alpha^2+\frac{\sigma^2}{|\mathbb T_{r}(t^*)|}}$.
\qed

\subsection{Proof of Theorem~\ref{Them_Uniform_Bound}}
We first derive the Lipschitz constant bounds of the posterior mean function $m_{\Delta}$ and the posterior variance function $\sigma^2_{\Delta}$. 
Consider the two predictions $m_{\Delta}(t^*)$ and $m_{\Delta}(t^{+})$ at two testing time steps 
$t^*, t^{+}\in \mathbb T$.
Based on 
Proportion~\ref{Prop_Model_Update},
we have that 
the absolute difference between the posterior means evaluated at $t^*$ and $t^{+}$ is given by 
\begin{align}
\nonumber
    &\left|m_{\Delta}(t^*)-m_{\Delta}(t^{+})\right| \\ \nonumber
    &= \|(K(t^*,T)-K(t^{+},T))\underbrace{(K(T,T)+\sigma^2I_n)^{-1}\Delta}_{\Psi}\|  \\ 
    &\leq L_k\sqrt{n}\|\Psi\|\left|t^{*}-t^{+}\right|=\frac{\alpha^2}{\beta e^{1/2}}\sqrt{n}\|\Psi\|\left|t^{*}-t^{+}\right|, \label{Eq_Mean_Lip}
\end{align}
where we use the result from~Lemma~\ref{Lem_Lip} that the squared exponential kernel is Lipschitz continuous in its first argument with constant $L_k=\frac{\alpha^2}{\beta e^{1/2}}$. 
Thus, the posterior mean function $m_{\Delta}$ 
at testing locations $t^*, t^{+}\in \mathbb T$ 
is Lipschitz continuous with the Lipschitz constant $L_m = \frac{\alpha^2}{\beta e^{1/2}}\sqrt{n}\|\Psi\|$.

Next, we derive the bound 
on the prediction variance function
$\sigma^2_{\Delta}$.
Applying the Cauchy-Schwarz inequality to the absolute value of the difference of posterior variances at the testing time steps $t^*$ and $t^{+}$, we have that
$\left| \sigma^2_{\Delta}(t^{*}) -\sigma^2_{\Delta}(t^{+}) \right|
   \leq\left\|K(t^*,T)-K(t^{+},T) \right\| 
   \times \left\|(K(T,T)+\sigma^2I_n)^{-1}\right\| 
   \times \|K(t^*,T)+K(t^{+},T)\|$, 
where we use the fact that $k(t^*,t^*)-k(t^{+},t^{+}) = 0$.
Again using Lemma~\ref{Lem_Lip} gives
$\left\|K(t^*,T)-K(t^{+},T) \right\|\leq \sqrt{n}\frac{\alpha^2}{\beta e^{1/2}}\left|t^{*}-t^{+}\right|$.  
In addition, by the definition of the  squared exponential kernel, we have 
\begin{align}
&\left\|K(t^*,T)+K(t^{+},T) \right\|\leq 2\sqrt{n}\underset{t^*, t^{+}\in \mathbb T}{\max} k(t^*,t^{+})\\
&=2\sqrt{n}\alpha^2 \exp \left(-\frac{\underset{t^*, t^{+}\in \mathbb T}{\min}{(t^*-t^{+})^2}}{2\beta^2}\right)=2\sqrt{n}\alpha^2.
\label{Eq_Kernel_Lip}
\end{align}
Hence,  the Lipschitz constant $L_{\sigma^2}$ of the posterior variance function $\sigma_{\Delta}^2$
is given by 
\begin{multline}
\left|\sigma^2_{\Delta}(t^{*})-\sigma^2_{\Delta}(t^{+}) \right| 
\\ \leq 
\underbrace{\frac{2n\alpha^4}{\beta e^{1/2}} 
    \left\|(K(T,T)+\sigma^2I_n)^{-1}\right\|}_{ L_{\sigma^2}} \left|t^{*}-t^{+}\right|.
\end{multline}
Then, we transfer the Lipschitz continuity of the posterior variance to the posterior standard deviation through the equation 
$|\sigma_{\Delta}(t^{*})-\sigma_{\Delta}(t^{+})|\leq \sqrt{\left|\sigma^2_{\Delta}(t^{*})-\sigma^2_{\Delta}(t^{+})\right|}$, due to the fact that standard deviations are nonnegative.  
Therefore,
we can bound the difference of the posterior standard deviation at two testing time steps 
through
\begin{align}
\left|\sigma_{\Delta}(t^{+})-\sigma_{\Delta}(t^{*})\right| 
\leq \sqrt{L_{\sigma^2}}\sqrt{\left|t^{+}-t^{*}\right|}. \label{Eq_SD_Lip}
\end{align}

Based on~\eqref{Eq_Mean_Lip}-\eqref{Eq_SD_Lip}, we show the high probability error bound of the posterior mean.  
Recall that we define the time interval of interest $\mathbb T$ such that the training time steps $T\subset \mathbb T$  and the testing time steps $t^*, t^{+} \in\mathbb T$.
Further, the time interval $\mathbb{T}$ is covered by $M(\tau, T)$ intervals 
of the form 
$\underline{\mathbb{T}}_{\tau}(t')$ for all $t'\in \mathbb{M}$.
We prove the high probability error bound by
exploiting the fact that if there exists  $t'\in \mathbb{M}$ such that
$t^+\in\underline{\mathbb{T}}_{\tau}(t')$,
and 
$\underset{t^*\in\mathbb{T}}{\max} \ \underset{t^{+}\in \underline{\mathbb{T}}_{\tau}(t')}{\min}\left|t^{*}-t^{+}\right|\leq \tau$, then 
it holds with a probability of at least 
$1-M(\tau, \mathbb{T})e^{-\gamma(\tau)/2}$
that
\begin{equation}
\left|\bar{\Delta}(t^*)-m_{\Delta}(t^*)\right|\leq\sqrt{\gamma(\tau)}\sigma_{\Delta}(t^*),
\end{equation}
for all $t^*\in\underline{\mathbb{T}}_{\tau}(t')$~\cite[Equations (25) and (26)]{srinivas2012information}.
By
choosing $\gamma(\tau)=2\log\left(\frac{M(\tau, \mathbb{T})}{\delta}\right)$, we have that
\begin{equation*}
    \left| \bar{\Delta}(t^*)-m_{\Delta}(t^*) \right| \leq \sqrt{2\log(\frac{M(\tau, \mathbb{T})}{\delta})}\sigma_{\Delta}(t^*),
\end{equation*}
for all  $t^*\in\underline{\mathbb{T}}_{\tau}(t')$ holds with a probability of at least $1-\delta$ for $\delta\in(0,1)$. 
 
Due to the continuity of  the functions  $\bar{\Delta}$,
$m_{\Delta}$ and $\sigma_{\Delta}^2$, we have that
$\underset{t^{+}\in \mathbb{T}_\tau(t')}{\min}\left|\bar{\Delta}(t^{*})-\bar{\Delta}(t^{+})\right|\leq \tau L_{\bar\Delta}$, for all $t^{*}\in \mathbb{T}$,
where $L_{\bar\Delta}$ is the Lipschitz constant of the function $\bar{\Delta}$. Based on~\eqref{Eq_Mean_Lip}, the prediction means between two time steps $t^{+}$ and $t^{*}$ satisfy
$\underset{t^{+}\in \underline{\mathbb{T}}_\tau(t')}{\min}\left|m (t^{*})-m_{\Delta}(t^{+})\right|\leq \tau L_m$, for all $t^{*}\in\mathbb{T}$,
where $L_m$ is the Lipschitz constant of the posterior mean function $m_{\Delta}$. In addition, according to~\eqref{Eq_SD_Lip}, the predicted standard deviation at two testing locations is given by
$\underset{t^{+}\in \underline{\mathbb{T}}_\tau(t')}{\min}\left|\sigma_{\Delta}(t^{+})-\sigma_{\Delta}(t^{*})\right|\leq  \sqrt{L_{\sigma^2}\tau}$, for all $ t^*\in\mathbb T$.
Based on these conditions and~\cite[Equations (27)-(33)]{lederer2019uniform}, 
we obtain that
\begin{equation}
p\left(\left| \bar{\Delta}(t^*)-m_{\Delta}(t^*)\right|\leq \sqrt{\gamma(\tau)}\sigma_{\Delta}(t^*)+\xi(\tau) \right)\geq 1-\delta,
\end{equation}
for all $t^*\in \mathbb{T}$, where $\gamma(\tau) = 2\log\left(\frac{M(\tau, \mathbb{T})}{\delta}\right)$, and 
$\xi(\tau) = (L_{\bar\Delta} +L_m)\tau + \sqrt{\gamma(\tau)L_{\sigma^2}\tau}$,
where $M(\tau, \mathbb T)$ is the covering
number of the interval $\mathbb T$
given the collection of points $\mathbb{M}$.

Recall that we use $\hat{\mathbb T}$ to represent the length of the interval $\mathbb T$.
By the definition of the covering
number of an interval, we have that 
$M(\tau, \mathbb T) = \frac{\hat{\mathbb T}}{2\tau}+1$.
Hence,
we have that $\gamma(\tau) = 2\log\left(\frac{\frac{\hat{\mathbb T}}{2\tau}+1}{\delta}\right)$, which completes the proof.
\qed
\end{document}